\documentclass{article}

\usepackage{microtype}
\usepackage{graphicx}
\usepackage{subfigure}
\usepackage{booktabs} 
\usepackage[flushleft]{threeparttable}
\usepackage{multirow}
\usepackage{xcolor}
\usepackage{amsmath}
\usepackage{bbm}
\usepackage{algorithm}
\usepackage{algorithmic}
\usepackage{amsfonts,amssymb}
\usepackage{booktabs,tabulary}
\usepackage{hyperref}
\usepackage{xspace}
\usepackage{comment}

\usepackage[accepted]{icml2020}

\usepackage{times}
\usepackage{graphicx} % more modern
\usepackage{subfigure} 

\usepackage{algorithm}
\usepackage{algorithmic}

\usepackage{helvet}
\usepackage{courier}

\usepackage{makecell}
\usepackage{graphicx}
\usepackage{subfigure}
\usepackage{color}
\usepackage{amsfonts}
\usepackage{amsmath}
\usepackage{amssymb}

\def\mL{{\mathcal L}}

\DeclareMathAlphabet\mathbfcal{OMS}{cmsy}{b}{n}
\def\0{{\bf 0}}
\def\1{{\bf 1}}

\newcommand{\ie}{\textit{i}.\textit{e}.}
\newcommand{\eg}{\textit{e}.\textit{g}.}
\def\etc{\emph{etc}.}

\usepackage{times}
\usepackage{epsfig}
\usepackage{graphicx}
\usepackage{amsmath}
\usepackage{amssymb}

\usepackage{subfigure}
\usepackage{booktabs}
\usepackage{multirow}

\usepackage{color}
\usepackage{capt-of}

\usepackage{comment}
\usepackage{array}

\usepackage{threeparttable}

\usepackage{amssymb}
\usepackage{pifont}

\usepackage{diagbox}

\newcommand{\cnas}{CNAS\xspace}
\newcommand{\cnasnode}{CNAS-Node\xspace}
\newcommand{\fixnas}{Fixed-NAS\xspace}

\newcommand{\eat}[1]{}

\def\mytitle{Breaking the Curse of Space Explosion: Towards Efficient NAS \\ with Curriculum Search
}

\icmltitlerunning{Breaking the Curse of Space Explosion: Towards Efficient NAS with Curriculum Search}

\begin{document}

\twocolumn[
\icmltitle{\mytitle}

\icmlsetsymbol{equal}{*}

\begin{icmlauthorlist}
\icmlauthor{Yong Guo}{equal,scut}
\icmlauthor{Yaofo Chen}{equal,scut,pzlab}
\icmlauthor{Yin Zheng}{equal,weixin}
\icmlauthor{Peilin Zhao}{equal,ailab}
\icmlauthor{Jian Chen}{scut,klab}
\icmlauthor{Junzhou Huang}{ailab}
\icmlauthor{Mingkui Tan}{scut}
\end{icmlauthorlist}

% Guangdong Key Laboratory of Big Data Analysis and Processing

\icmlaffiliation{scut}{School of Software Engineering, South China University of Technology}
\icmlaffiliation{weixin}{Weixin Group, Tencent}
\icmlaffiliation{ailab}{Tencent AI Lab, Tencent}
% \icmlaffiliation{uta}{University of Texas at Arlington}
\icmlaffiliation{klab}{Guangdong Key Laboratory of Big Data Analysis and Processing}
\icmlaffiliation{pzlab}{Pazhou Laboratory}

\icmlcorrespondingauthor{Mingkui Tan}{mingkuitan@scut.edu.cn}
\icmlcorrespondingauthor{Jian Chen}{ellachen@scut.edu.cn}

\icmlkeywords{Machine Learning, ICML}

\vskip 0.3in
]

\printAffiliationsAndNotice{\icmlEqualContribution. This work is done when Yong Guo works as an intern in Tencent AI Lab.} % 

\begin{abstract}
Neural architecture search (NAS) has become an important approach to automatically find effective architectures.
To cover all possible good architectures, we need to search in an extremely large search space with billions of candidate architectures. 
More critically, given a large search space, we may face a very challenging issue of space explosion. 
However, due to the limitation of computational resources, we can only sample a very small proportion of the architectures, which provides insufficient information for the training. 
As a result, existing methods may often produce sub-optimal architectures. To alleviate this issue, 
we propose a curriculum search method that starts from a small search space and gradually incorporates the learned knowledge to guide the search in a large space.
With the proposed search strategy, our Curriculum Neural Architecture Search (CNAS) method significantly improves the search efficiency and finds better architectures than existing NAS methods.
Extensive experiments on CIFAR-10 and ImageNet demonstrate the 
effectiveness of the proposed method.
\end{abstract}

\section{Introduction}

Deep neural networks (DNNs) have been producing state-of-the-art results in many challenging tasks including image classification~\cite{krizhevsky2012imagenet,liu2020discrimination,zheng2015neural,guo2020multi}, 
semantic segmentation~\cite{DBLP:journals/pami/ShelhamerLD17, DBLP:journals/pami/ChenPKMY18,huang2019ccnet}, and many other areas~\cite{zheng2016deep,zheng2016neural,Jiang2017,chen2019generalized,zeng2019graph,hosseini2019deep,guo2020closed,pmlr-v80-cao18a}.
% object detection~\cite{ren2015faster,DBLP:conf/cvpr/RedmonDGF16,NIPS2019_8890}. 
Besides designing DNNs manually, there is a growing interest in automatically designing effective architectures by neural architecture search (NAS) methods~\cite{zoph2016neural,pham2018efficient}. 
Existing studies show that the automatically searched architectures often outperform the manually designed ones in both image classification tasks and language modeling tasks~\cite{zoph2018learning,lian2019towards}.

\begin{figure}[t]
	\centering
	\includegraphics[width=1\columnwidth]{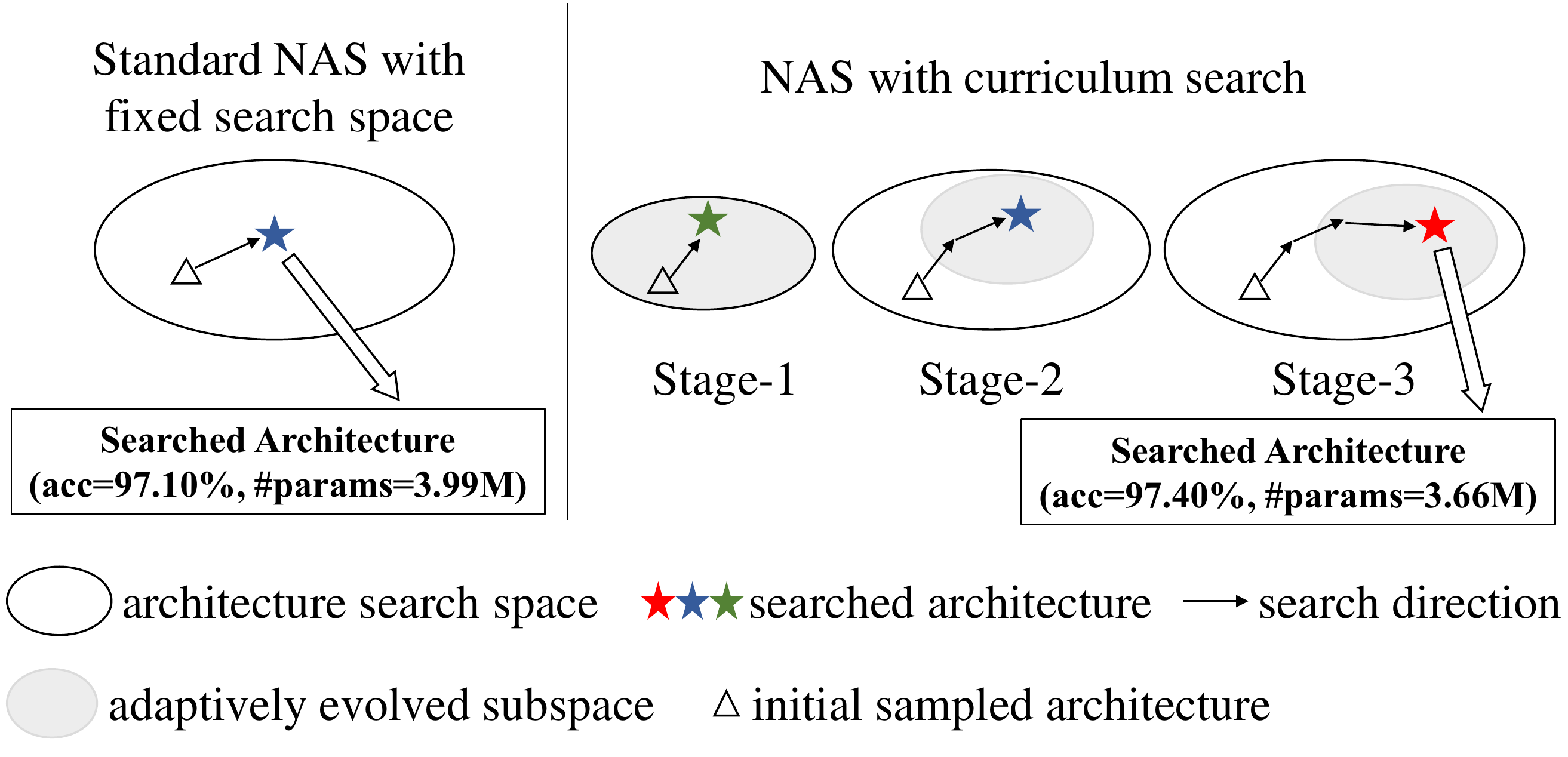}
	\caption{
	    Comparisons of the search process between {standard NAS methods and our proposed curriculum NAS (CNAS) method.} By gradually refining the candidate subspace, \cnas performs much accurate architecture sampling, thus can significantly improve the search efficiency and find better architectures than standard NAS methods.
	}
	\label{fig:cnas_scheme}
\end{figure}

NAS methods seek to search for an optimal architecture in a predefined search space.
To this end, existing methods have to explore the whole search space by sampling sufficiently many architectures.
However, the search space is often extremely large (\eg, billions of candidate architectures~\cite{pham2018efficient}), resulting in the space explosion issue.
To be specific, due to the limitation of computational resources, we can only sample a very small proportion of the architectures from the search space.
Thus, we can only receive very limited information to guide the search process.
As a result, NAS models become hard to train and often {find} sub-optimal architectures. 

To alleviate the space explosion issue, we seek to improve the search by conducting more accurate sampling, \ie, exploring the subspace that contains potentially better architectures.
In this way, given limited resources to conduct sampling, we can find good architectures with relatively high probability and thus improve the search performance.
However, how to train NAS models to perform more accurate sampling is still unknown.

To address the above issues, we propose a novel curriculum search method to improve the search performance.
Specifically, we first conduct the search in a relatively small space where sufficient exploration is possible.
Then, we gradually enlarge the search space and use the previously learned knowledge to make the sampling more accurate in the larger search space. 
The key idea follows the basic concept of curriculum learning that humans and animals can learn much better when they gradually learn new knowledge~\cite{bengio2009curriculum,gulccehre2016knowledge}.

Equipped with the curriculum search, we propose a Curriculum Neural Architecture Search (\cnas) method.
Our \cnas enlarges the search space by gradually increasing the number of candidate operations and exploits the previously learned knowledge to achieve accurate sampling.
As shown in Figure~\ref{fig:cnas_scheme}, once we have found some good architectures in a small search space and gradually enlarge the space, it becomes more likely to find a candidate subspace (grey circle) that shares some common knowledge with the previous one but contains potentially better architectures.
When we consider multiple stages to perform architecture search, the subspace would adaptively evolve along the search process.
Based on the adaptively evolved subspace, \cnas is able to conduct more accurate sampling to find better architectures in a large search space.
Furthermore, to improve the training stability when we introduce a previously unseen operation, we propose an operation warmup strategy to make all the operations relatively fair when sampling architectures.
Extensive experiments demonstrate the superiority of our \cnas over existing methods.

Our contributions are summarized as follows:
\begin{itemize}
    \item We propose a novel Curriculum Neural Architecture Search (CNAS) method to alleviate the training difficulties of the NAS problem incurred by the extremely large search space.
    To this end, we break the original NAS problem into a series of simpler problems and train the controller model in a progressive manner.
    \item We propose a curriculum search method that gradually incorporates the knowledge learned from a small search space. 
    To this end, we start from a search space with one operation and gradually add new operations. Thus, the previously learned knowledge about how to use a specific operation can be effectively preserved.  
    \item Extensive experiments on several benchmark data sets show that the architectures found by our \cnas significantly outperform the architectures obtained by state-of-the-art NAS methods. 
\end{itemize}

\section{Related Work}

\textbf{Neural architecture search.}
In the past few years, neural architecture search (NAS) has attracted considerable attention to automatically design effective architectures.
\citet{zoph2016neural} first propose to learn a controller {for} an optimal configuration of each convolution layer.
However, it performs search for an entire network, leading to extremely large search space and very high search cost.
To reduce the search space, NASNet~\cite{zoph2018learning,pham2018efficient} proposes to search for the optimal neural cell rather than the whole network.
Related to our method, \citet{liu2018progressive} propose a PNAS method that gradually enlarges the search space and performs architecture search in a progressive manner.
Specifically, PNAS picks the top $K$ architectures in each stage and gradually adds nodes to progressively enlarge the search space.
However, there are several limitations with PNAS. First, the unselected architectures as well as all the architectures that are obtained by adding additional nodes would be ignored. As a result, it greatly limit the possible search space to find good architectures in the next stage. Second, PNAS has to train a large number of architectures until convergence to learn a performance predictor, resulting in extremely high search cost (\ie, 255 GPU days).

\textbf{Curriculum learning.}
\citet{bengio2009curriculum} propose a new learning strategy called curriculum learning, which improves the generalization ability of model and accelerates the convergence of the training process.
Recently, many efforts have been made to design effective curriculum learning methods~\cite{Wang_2019_ICCV,DBLP:journals/ijcv/ZhangHZM19}. 
\citet{kumar2010self} propose a self-paced learning algorithm that selects training samples in a meaningful order. 
\citet{khan2011humans} provide evidence about the consistency between curriculum learning and the learning principles of human.
\citet{bengio2013representation} further present insightful explorations for the rationality of curriculum learning.
\citet{matiisen2019teacher} build a framework that produces Teacher-Student curriculum learning. 
Moreover, curriculum learning has been applied to NAS~\cite{cheng2018instanas,zoph2016neural}. ~\citet{cheng2018instanas} propose an InstaNAS method that uses a dynamic reward function to gradually increase task difficulty, and~\citet{zoph2016neural} propose to gradually increase the number of layers.
Unlike these methods, we construct simpler problems from a new perspective of search space in the context of the NAS problem.

\section{Preliminary} \label{section:perliminary}
	
For convenience, we revisit the Neural Architecture Search (NAS) problem. 
Reinforcement Learning (RL) based NAS methods seek to learn a controller to produce candidate architectures.
Let $\theta$ be the trainable parameters and $\Omega$ be the search space. {A} controller can produce a candidate architecture $\alpha$ as follows:
\begin{equation}
\alpha \sim \pi(\alpha; \theta,\Omega),
\end{equation}
where $\pi$ is {a} policy learned by the controller, \ie, a distribution of candidate architectures. 
In practice, the policy learned by the controller can be either the discrete form (\eg, NAS~\cite{zoph2016neural} and ENAS~\cite{pham2018efficient}), or the differentiable form like DARTS~\cite{liu2018darts}.
	
To measure the performance of an architecture, we have to train {a} model {built} with the architecture {until} convergence. Let $\mL$ be the loss function on training data. Given an architecture $\alpha$, the optimal parameters $w^*(\alpha)$ can be obtained by $w^*(\alpha)=\arg\min_{w}\mathcal{L}\left(\alpha,w\right)$.
Then, one can measure the performance of $\alpha$ by some metric {$R\left(\alpha, w^*(\alpha)\right)$}, \eg, the accuracy on validation data.
Following the settings of NAS~\cite{zoph2016neural} and ENAS~\cite{pham2018efficient}, we use Reinforcement Learning (RL) and take the performance $R(\cdot)$ as the ``reward'' to train the controller model.
	
The goal of RL-based NAS is to find {an} optimal policy by maximizing the expectation of the reward $R(\alpha, w^*(\alpha))$. 
Thus, the NAS problem can be formulated as a bilevel optimization problem:
\begin{equation} \label{eq:nas-general}
\begin{aligned}
\max_{\theta}~&\mathbb{E}_{\alpha\sim \pi(\alpha;\theta,\Omega)}\mathcal{R}\left(\alpha,w^{*}(\alpha)\right)\\
\text{s.t. }&w^*(\alpha)=\arg\min_{w}\mathcal{L}\left(\alpha,w\right).
\end{aligned}
\end{equation}
To solve this problem, one can update $\theta$ and $w$ in an alternative manner~\cite{zoph2016neural,pham2018efficient}.

\section{Curriculum Neural Architecture Search}
    
For NAS methods, the huge search space is often the main bottleneck to the architecture search performance. 
In general, a small space often leads to sub-optimal architectures with inferior performance, but a large search space would incur severe space explosion issue and may even make the learning task infeasible. Specifically, since we can only explore a very small proportion of architectures to train the controller, it is very difficult to learn a NAS model to find good architectures in a large search space.
    
In this paper, we seek to improve NAS by alleviating the space explosion issue.
We first analyze the size of the search space. Then, we propose a curriculum search strategy to {break the original NAS problem into a series of simpler problems and then we solve each problem in a progressive manner.} For convenience, we call our method Curriculum Neural Architecture Search (\cnas). An illustrative comparisons between the standard NAS methods and the proposed \cnas can be found in Figure~\ref{fig:cnas_scheme}. 
    
\begin{figure}[t]
	\centering
	\includegraphics[width=1.0\columnwidth]{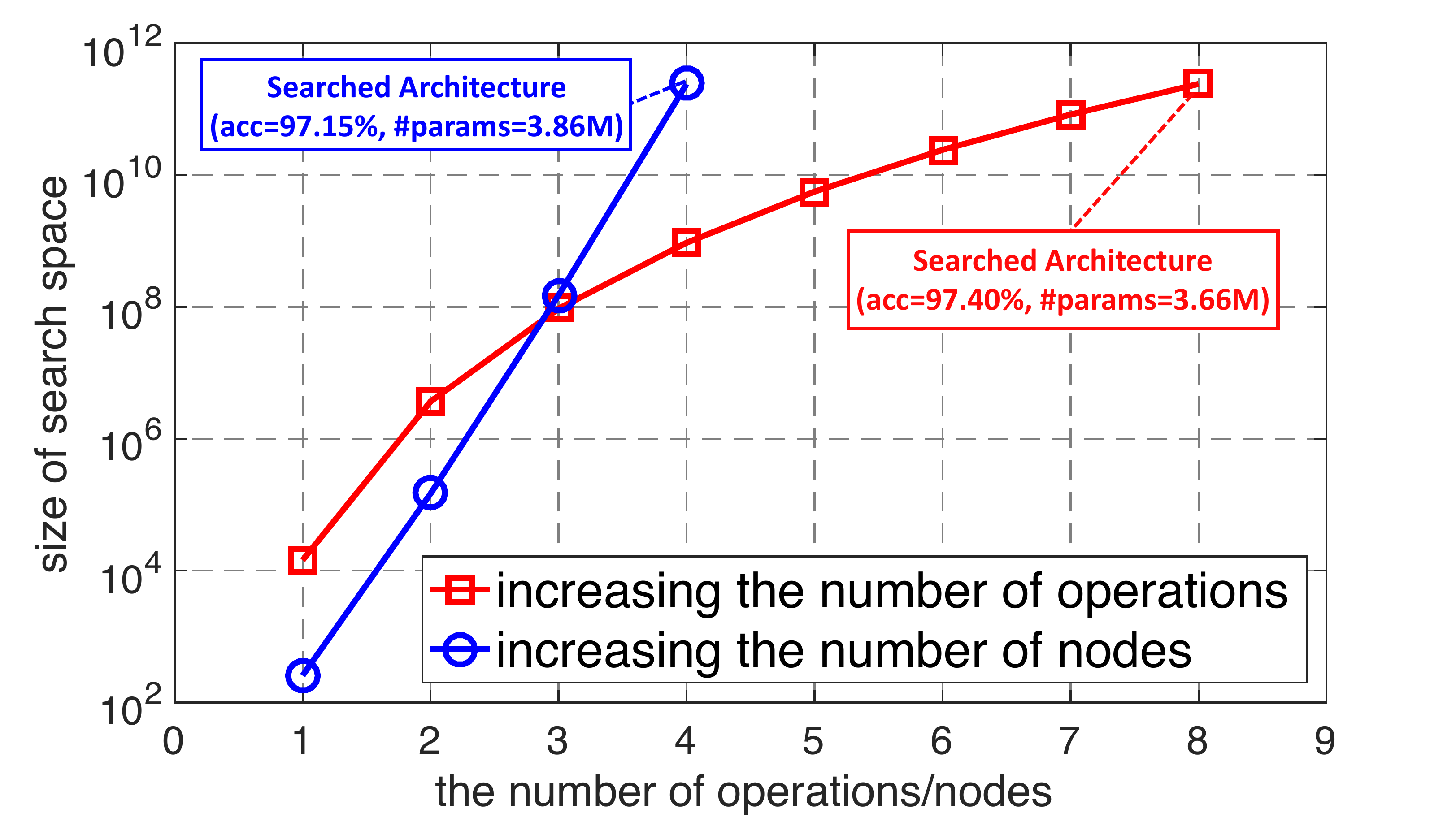}
	\caption{
	    {Comparisons of different search spaces on one cell architecture. Following~\cite{liu2018darts},} we consider a cell architecture with 4 intermediate nodes and 8 candidate operations. 
	}
	\label{fig:space_size}
\end{figure}
    
\subsection{Search Space Size Analysis}\label{sec:space}
	
We consider learning a generic computational cell because searching a good architecture for {an} entire network is very computationally expensive~\cite{zoph2018learning}.
In NAS methods~\cite{zoph2016neural,pham2018efficient,liu2018darts}, a cell-based architecture $\alpha$ can be represented by a directed acyclic graph (DAG),
\ie, $\alpha = (\mathcal{V}, \mathcal{E})$.
Here, $\mathcal{V}$ is the set of nodes that represent the feature maps in {a} neural network.
$\mathcal{E}$ is the set of the edges that represent some computational operations (\eg, convolution or pooling).
For convenience, we denote the numbers of nodes by $B$.
	
Following~\cite{liu2018darts}, a DAG contains two input nodes, $B-3$ intermediate nodes, and one output {node}. The input nodes denote the outputs of the nearest two cells in front of the current one. The output node concatenates the outputs of all the intermediates to produce a final output of the cell. In the DAG, each intermediate node also takes two previous nodes in this cell as inputs. 
In this sense, there are $2 \times (B - 3)$ edges in the DAG and we will determine which operation should be applied to each of them.

Given $B$ nodes and $K$ candidate operations, the size of the search space $\Omega$ can be computed by\footnote{{More analysis of search space size is put in supplementary.}}
\begin{equation}\label{eq:search_space}
    |\Omega| = K^{2(B-3)}\big((B-2)!\big)^2.
\end{equation}
From Eqn.~(\ref{eq:search_space}), the search space can be extremely large when we have a large $B$ or $K$.
For example, ENAS~\cite{pham2018efficient} has a search space of $|\Omega| \approx 5 {\times} 10^{12}$ with $B {=} 8$ and $K {=} 5$, and DARTS~\cite{liu2018darts} has a search space of $|\Omega| \approx 2 {\times} 10^{11}$ with $B {=} 7$ and $K {=} 8$.
In the extremely large search space, we can only sample a very limited number of architectures. As a result, the feedback/reward from the sampled architectures is insufficient, making it hard to train a good controller. As a result, the search process may severely suffer from the space explosion issue.

\begin{figure}[t]
	\centering
	\includegraphics[width=0.9\columnwidth]{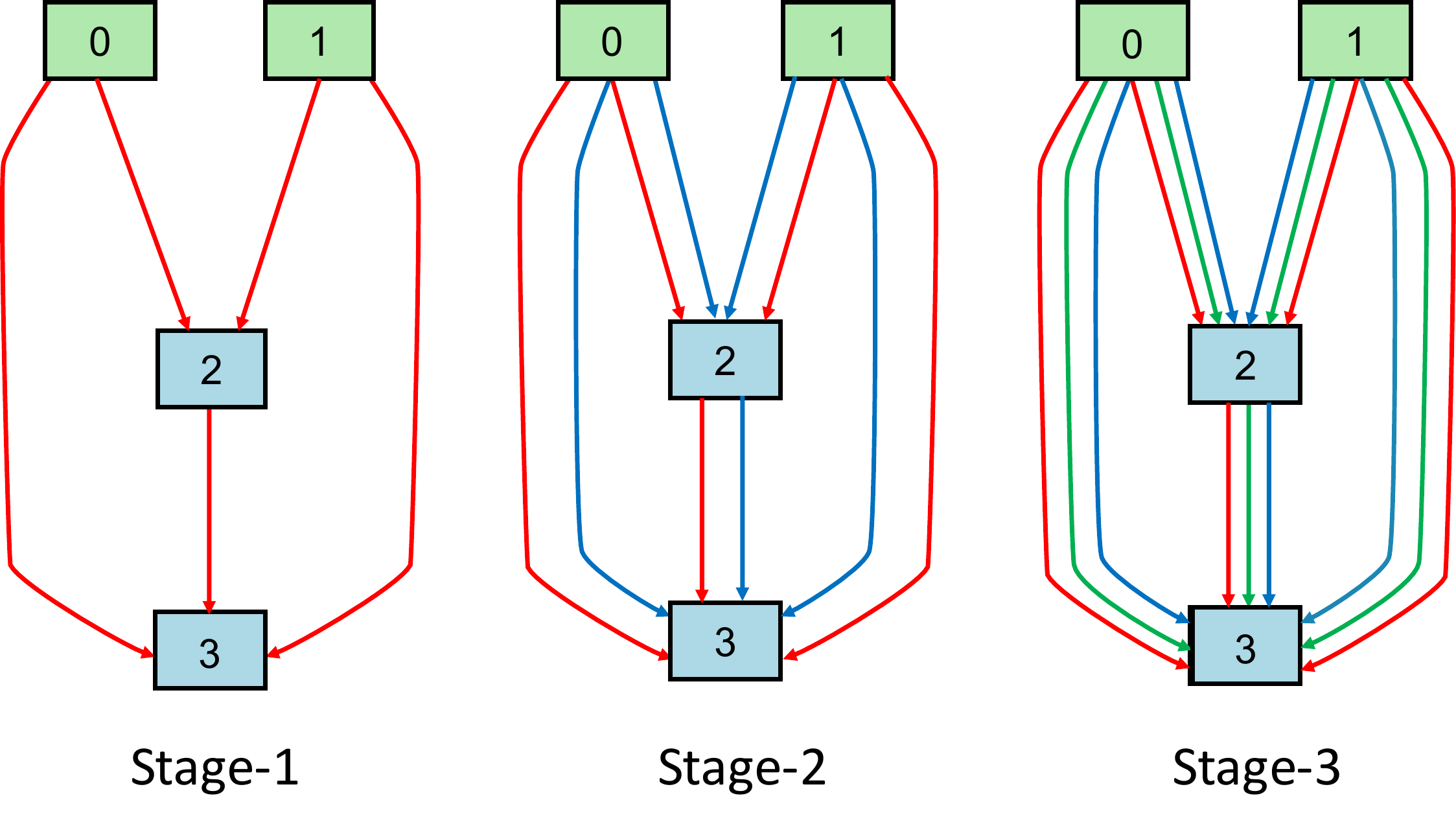}
	\caption{
        An overview of the search space used by \cnas. We show the candidate operations of the super network in different stages. The edges with different colors denote different operations. For simplicity, we omit the output node in this figure.
	}
	\label{fig:progressive_space}
\end{figure}

\begin{algorithm}[t]
	\caption{\small{Training method for CNAS.}}
    	\begin{algorithmic}[1]\small
    		\REQUIRE The operation sequence $O$, learning rate $\eta$, the number of the iterations for operation warmup $M$, the uniform  distribution of architectures $p(\cdot)$, the controller's policy $\pi(\cdot)$, super network parameters $w$, controller parameters $\theta$.
            \STATE {Initialize $w$ and $\theta$, $\Omega_0 = \O$.}
            \FOR{$i{=}1$ to $|O|$}
            \STATE Enlarge $\Omega_i$ by adding $O_i$ to the set of candidate operations;\\
            \STATE // \emph{Operation warmup}
            \FOR{$j{=}1$ to $M$}
                \STATE Sample $\alpha \sim p(\alpha ; \Omega_i)$;
                \STATE $w \gets w - \eta \nabla_{w} \mathcal{L}(\alpha,w)$;
            \ENDFOR
            \WHILE{not convergent}
        		\STATE // \emph{Update $\theta$ by maximizing the reward}
        		\FOR{each iteration on validation data}
                    \STATE Sample $\alpha \sim \pi(\alpha;\theta,\Omega_i)$;
                    \STATE Update the controller by ascending its gradient:
            		\STATE ~~~~$ \mathcal{R}(\alpha,w) \nabla_\theta\log \pi(\alpha;\theta,\Omega_i) {+} \lambda H(\pi(\cdot;\theta,\Omega_i)) $; \\
            	\ENDFOR
        		\STATE // \emph{Update $w$ by minimizing the training loss}
        		\FOR{each iteration on training data}
                    \STATE Sample $\alpha \sim \pi(\alpha;\theta,\Omega_i)$;
                    \STATE $w \gets w - \eta \nabla_{w} \mathcal{L}(\alpha,w)$. \\
                \ENDFOR
                \ENDWHILE
            \ENDFOR
    	\end{algorithmic}
		\label{alg:training}
\end{algorithm}
    
\subsection{NAS with Curriculum Search}

To alleviate the space explosion issue, 
we seek to improve the search process by providing more powerful information to improve the quality of architecture sampling. To this end,
we exploit the idea of curriculum learning that humans often learn much better when they gradually learn new knowledge/concepts.
Specifically, we propose to break the NAS problem into a series of simpler problems.
Since the size of the search space is an indicator of the difficulty level of the NAS problem,
we may change the size of search space to construct the problems with different difficulty levels. 
In this sense, we can cast the training of NAS models into a multi-stage process to gradually incorporate previously learned knowledge to find better architectures.
    
As mentioned in Section~\ref{sec:space}, the size of search space $|\Omega|$ depends on the number of nodes $B$ and the number of candidate operations $K$. In this sense, we can adjust either $B$ or $K$ to obtain the search spaces with different sizes. To find a better choice between these two methods, we compare the search space w.r.t. different $B$ and $K$. 
Following~\cite{pham2018efficient,liu2018darts}, we adopt a widely used setting with $B{=}7$ nodes (\ie, 4 intermediate nodes) and $K{=}8$ candidate operations. In this case, we investigate the effect of increasing the number of nodes and operations on the size of the search space in Figure~\ref{fig:space_size}.
From Figure~\ref{fig:space_size} and Eqn.~(\ref{eq:search_space}), increasing node would make the size of search space grow much faster than increasing operation.
As a result, increasing node would introduce a {non-}negligible gap between adjacent stages.
Thus, the training difficulty incurred by the extremely increased search space is still severe.
On the contrary, increasing operation from $1$ to $K$ provides a more smooth growth of search space, making progressive training possible (See the detailed comparisons in Section~\ref{exp:node_vs_operation}).
    
Thus, we seek to enlarge the search space by gradually increasing the number of candidate operations.
Specifically, we start from the search space with a single operation and then add a new operation to the set of candidate operations in each stage. 
To accelerate the search process, we adopt the parameter sharing~\cite{pham2018efficient} technique that makes all the child networks share their parameters in a super network. 
For clarity, we show the super network with the progressively growing search space in Figure~\ref{fig:progressive_space}.
Without loss of generality, we add the operations in {a} random order (See discussions about the order in Section~\ref{exp:order}). 

\noindent \textbf{Curriculum training scheme.}
Based on the curriculum search strategy, we can obtain a series of problems with different difficulty levels.
However, how to effectively solve these problems to improve the training of the controller model still remains a question.
To address this issue, we propose a curriculum training algorithm and show the details in Algorithm~\ref{alg:training}.
Specifically, we progressively train the controller to solve the problems with different search spaces.
During the training, we gradually increase the number of operations from $1$ to $K$. Thus, the whole training process can be divided into $K$ stages. 
To encourage the diversity when sampling architectures, we introduce an entropy term into the objective.
Let $\Omega_i$ be the search space of the $i$-th stage. The training objective in this stage can be written as
\begin{equation} \label{eq:cnas_stage}
	\begin{aligned}
	\max_{\theta}~&\mathbb{E}_{\alpha\sim \pi(\cdot;\theta,\Omega_i)} \left[ \mathcal{R}\left(\alpha,w^{*}(\alpha)\right) \right] + \lambda H\left(\pi \left (\cdot;\theta,\Omega_i \right) \right) \\
	&~~~~~~~\text{s.t. }w^*(\alpha)=\arg\min_{w}\mathcal{L}\left(\alpha,w\right),
	\end{aligned}
\end{equation}
where $\pi(\cdot;\theta,\Omega_i)$ denotes the learned policy \emph{w.r.t.} $\Omega_i$, {$H(\cdot)$ evaluates the entropy of the policy, and $\lambda$ controls the strength of the entropy regularization term. 
Note that $\pi(\alpha;\theta,\Omega_i)$ denotes the probability to sample some architecture $\alpha$ from the policy/distribution $\pi(\cdot;\theta,\Omega_i)$. 
This entropy term enables CNAS to explore the unseen areas of previous search stages and thus escape from local optima.}
    
\textbf{Inferring architectures}.
Once we obtain a good controller model, we can use it to infer good architectures.
Given $K$ candidate operations, we take the learned policy $\pi(\cdot; \theta, \Omega_K)$ obtained in the final stage, \ie, the one with the largest search space, as the final policy to sample architectures.
Following~\cite{zoph2016neural,pham2018efficient}, we first sample 10 architectures and then select the architecture with the highest validation accuracy. 

\subsection{Operation Warmup}

In \cnas, we gradually add new operations into the set of candidate operations.
However, the corresponding parameters of the new operation are randomly initialized while the old operations have been extensively trained, leading to severe unfairness issues among operations~\cite{chu2019fairnas}.
As a result, the architectures with the new operation often yield very low rewards, making the controller tend not to choose it in the following training process.

To address this issue, we propose an effective operation warmup method.
Specifically, when we add a new operation, we fix the controller model and only train the parameters of the super network.
To improve the fairness of operations, we uniformly sample candidate architectures to train each operation with equal probability~\cite{chu2019fairnas}.
In this way, the candidate architectures with the newly added operation achieve comparable performance with the existing ones.
With the operation warmup method, we make the search process more stable and obtain significantly better search performance (See results in Section~\ref{exp:warmup}).

\section{More Discussions on CNAS}

In this section, we conduct further analysis of the proposed \cnas method. We first investigate our advantages over the existing NAS methods. Then, we discuss the differences between \cnas and a related work PNAS.

\subsection{Advantages of \cnas over the standard NAS}
The major advantage lies in the proposed curriculum search strategy.
Specifically, \cnas {trains} the controller in a small search space in the early stage.
Compared with the large search space, we can easily obtain a good controller since we can sufficiently explore the small search space (\eg, $|\Omega| = 120$ when $K=1$).
In this case, we do not need to consider which operation should be chosen but learn {an} optimal cell topology (\ie, node connection method).
When we gradually increase $K$, \cnas only needs to learn the new concept (\ie, the new operation) to fit the larger search space.
More critically, we can take the previously learned knowledge about which cell topology is good and explore the subspace that shares similar topology in the larger space.
As a result, it is more likely to find better architectures compared with the standard NAS method searched in a fixed search space (See results in Section~\ref{exp:node_vs_operation}).

\subsection{Differences from PNAS} 
A related work PNAS~\cite{liu2018progressive} also conducts architecture search in a progressive manner.
However, there exist several major differences between our method and PNAS. First, PNAS gradually increases the number of nodes to conduct a progressive search.
However, we analyze the size of the search space and propose to gradually enlarge the search space by introducing additional operations. 
Second, PNAS exploits a heuristic search method that periodically removes a large number of possible  architectures from the search space and thus limits the exploration ability. 
However, \cnas performs architecture search in the original search space specified by each stage, making it possible to find potentially better architectures. 
Third, PNAS has to train a large number of architectures until convergence to learn a performance predictor, resulting in extremely high search cost (\eg, 255 GPU days). However, \cnas exploits the weight sharing technique~\cite{pham2018efficient} and yields significantly lower search cost (See Table~\ref{tab:cifar-10}).

\section{Experiments} \label{section:experiments}

We apply the proposed \cnas to train the controller model on CIFAR-10~\cite{krizhevsky2009learning}. 
Then, we evaluate the searched architectures on CIFAR-10 and ImageNet~\cite{deng2009imagenet}.
All the implementations are based on PyTorch.\footnote{The code is available at \href{https://github.com/guoyongcs/CNAS}{https://github.com/guoyongcs/CNAS}.}
We organize the experiments as follows.
First, to demonstrate the effectiveness of our proposed \cnas, we compare the performance of the proposed \cnas with two related variants.
Second, we compare the performance of the architectures searched by \cnas with state-of-the-art image {classification methods} on CIFAR-10 and ImageNet.

\begin{figure}[t]
	\centering
	\includegraphics[width=1.05\columnwidth]{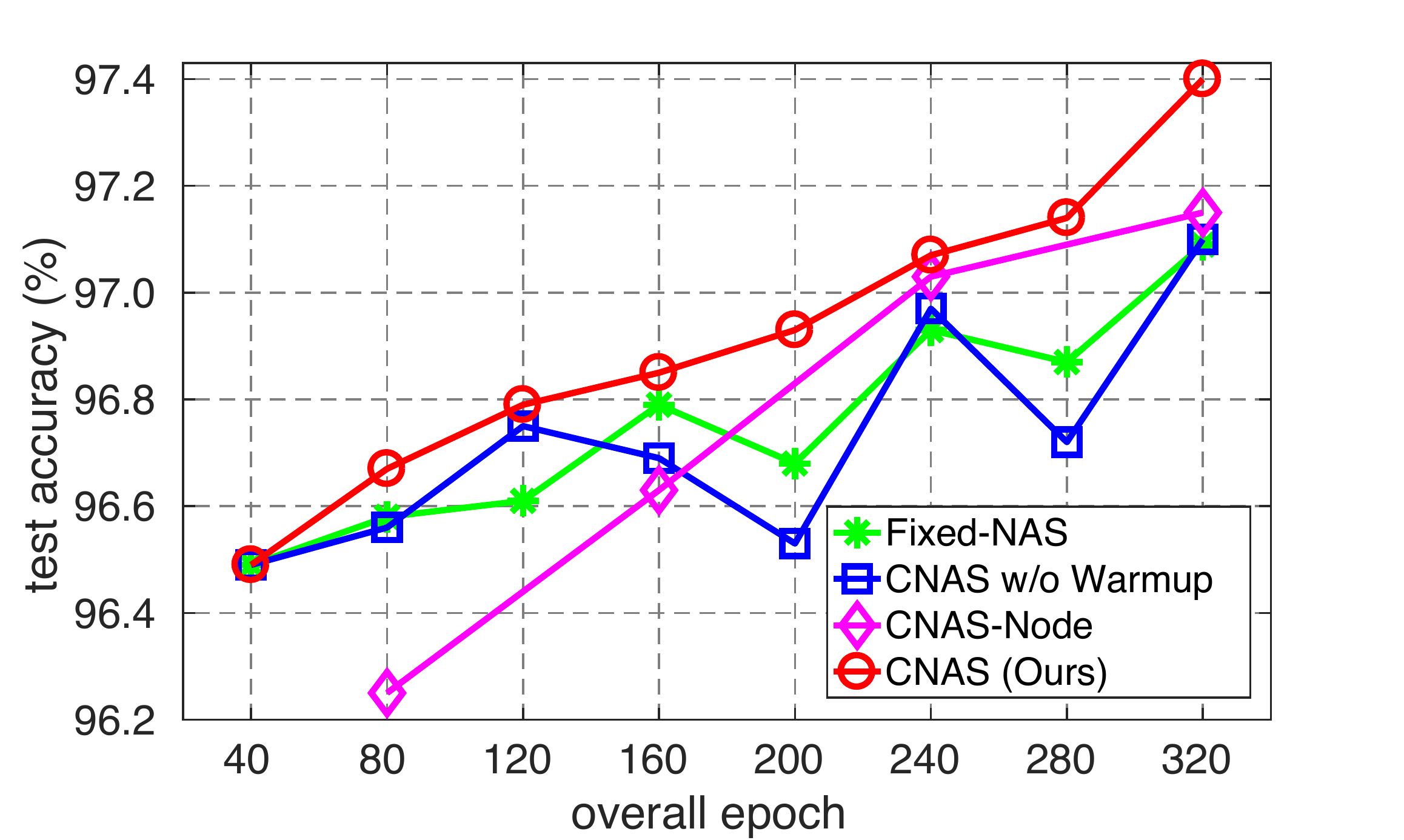}
	\caption{
	    Performance comparisons of the architectures obtained by different methods during the search process. All the models are evaluated on the test set of CIFAR-10. Each point indicates the average performance of the architectures searched over 5 independent experiments in different stages.
	}
	\label{fig:curriculum}
\end{figure}

\textbf{Compared methods}.
To investigate the effect of the proposed curriculum search strategy, we investigate and compare the following methods:
1) \textbf{\fixnas}: For each stage of \cnas, we keep the search space fixed and train a controller from scratch.
Following the settings in~\cite{liu2018darts}, we set the number of the nodes $B$ to $7$ (\ie, 4 intermediate nodes) and the number of candidate operations $K$ to $8$.
2) \textbf{\cnas}: We train the controller in a growing search space by gradually adding new operations while keeping $B$ unchanged.
3) \textbf{\cnasnode}: By fixing $K$, we gradually increase $B$ from 1 to 4.
We also compare the proposed \cnas with several state-of-the-art image classification methods, such as NASNet~\cite{zoph2018learning}, AmoebaNet~\cite{real2018regularized}, PNAS~\cite{liu2018progressive}, ENAS~\cite{pham2018efficient}, DARTS~\cite{liu2018darts}, \etc

\begin{table*}[t]
	\centering
	\caption{Comparisons with state-of-the-art models on CIFAR-10. We report the mean and standard deviation of the test accuracy over 10 independent experiments for different models.}
	\label{tab:cifar-10}
	\resizebox{0.91\textwidth}{!}
    {
	\begin{tabular}{cccc}
		\hline
		Architecture & Test Accuracy (\%)& Params (M) & Search Cost (GPU days)\\
		\hline
		DenseNet-BC~\cite{huang2017densely}&96.54&25.6 &--\\
		PyramidNet-BC~\cite{han2017deep}&96.69&26.0 &--\\
		\hline
		Random search baseline &96.71 $\pm$ 0.15 &3.2 &--\\
		NASNet-A + cutout~\cite{zoph2018learning}&97.35&3.3 &1800\\
		NASNet-B~\cite{zoph2018learning}&96.27&2.6 &1800\\
		NASNet-C~\cite{zoph2018learning}&96.41&3.1 &1800\\
		AmoebaNet-A + cutout~\cite{real2018regularized}&96.66 $\pm$ 0.06&3.2 &3150\\
		AmoebaNet-B + cutout~\cite{real2018regularized}&96.63 $\pm$ 0.04&2.8 &3150\\
		DSO-NAS~\cite{zhang2018you} & 97.05 & 3.0 & 1\\
		Hierarchical Evo~\cite{liu2017hierarchical}&96.25 $\pm$ 0.12 &15.7 &300\\
		SNAS~\cite{xie2018snas}&97.02&2.9 &1.5\\
		ENAS + cutout~\cite{pham2018efficient}&97.11&4.6 &0.5\\
		NAONet~\cite{luo2018neural}&97.02&28.6 &200\\
		NAONet-WS~\cite{luo2018neural}&96.47&2.5 &0.3\\
		GHN~\cite{zhang2018graph} & 97.16 $\pm$ 0.07 & 5.7 & 0.8 \\
		PNAS + cutout~\cite{liu2018progressive} & 97.17 $\pm$ 0.07 &3.2 &225\\
		DARTS + cutout~\cite{liu2018darts} &97.24 $\pm$ 0.09 &3.4 &4\\
        CARS + cutout~\cite{yang2019cars} &97.38 &3.6 & 0.4 \\
		\hline
		\cnas + cutout & \textbf{97.40 $\pm$ 0.06} & 3.7 & 0.3 \\
		\hline		
	\end{tabular}
	}
\end{table*}

\begin{figure*}[t]
	\centering
	\subfigure[Normal cell.]
	{
		\label{fig:normal}
		\includegraphics[width = 1\columnwidth]{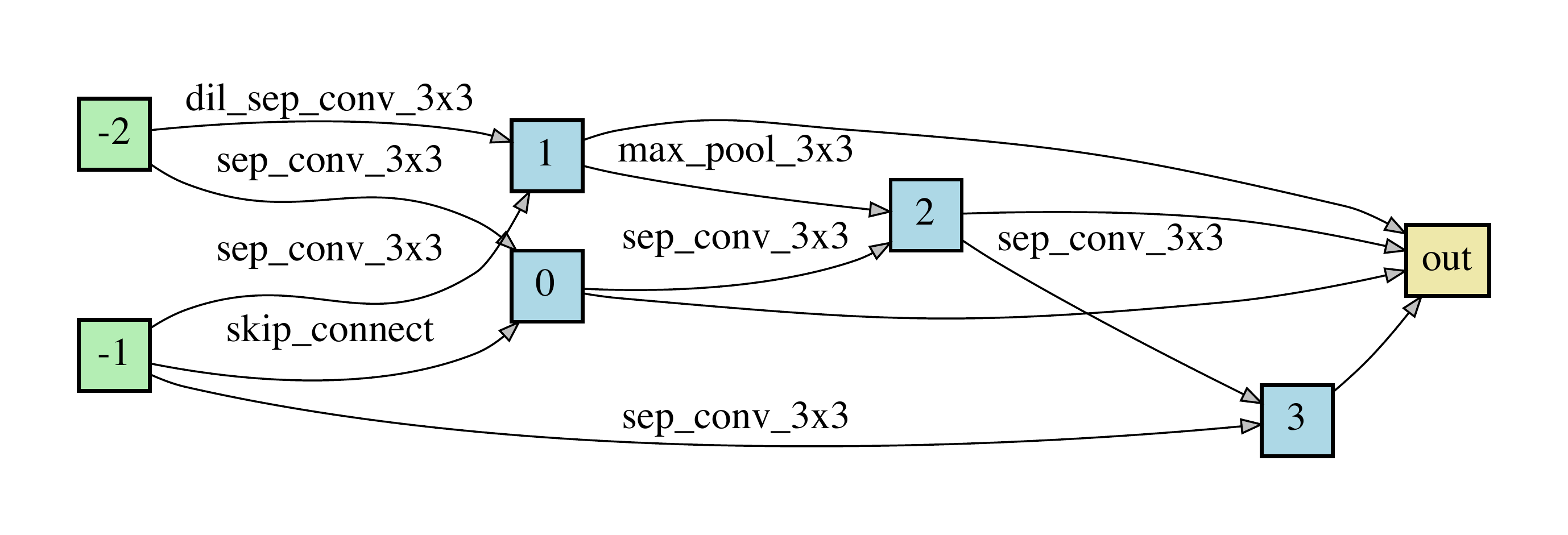}
	}
	\subfigure[Reduction cell.]
	{
		\label{fig:reduction}
		\includegraphics[width = 1\columnwidth]{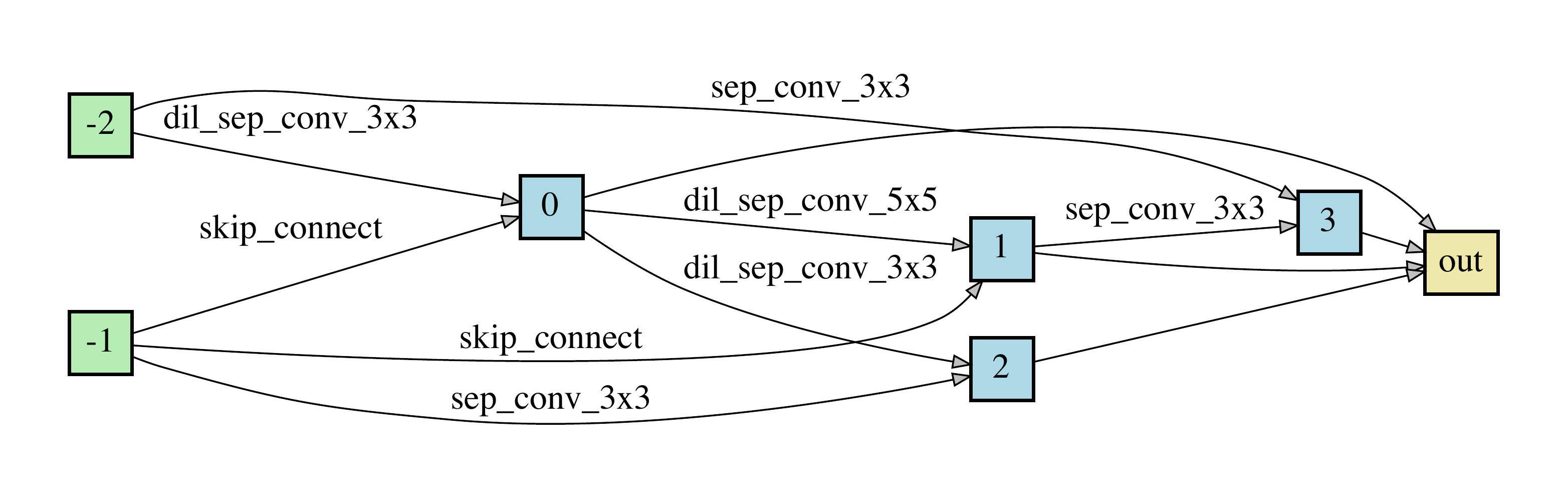}
	}
	\caption{The architecture of the convolutional cells found by \cnas. We conduct architecture search on CIFAR-10 and evaluate the architecture on both CIFAR-10 and ImageNet datasets.}
	\label{fig:cnn_arch}
\end{figure*}

\begin{table*}[t]
	\centering
	\caption{Comparisons with state-of-the-art image classifiers on ImageNet. ``-'' denotes the results that are not reported.}
	\label{tab:imagenet}
	\resizebox{0.88\textwidth}{!}
	{
    \begin{tabular}{cccccc}
    \hline
    \multicolumn{1}{c}{\multirow{2}[0]{*}{Architecture}} & \multicolumn{2}{c}{Test Accuracy (\%)} & \multicolumn{1}{c}{\multirow{2}[0]{*}{\#Params (M)}} & \multicolumn{1}{c}{\multirow{2}[0]{*}{\#MAdds (M)}} & \multicolumn{1}{c}{Search Cost} \\
    \cline{2-3}
          & \multicolumn{1}{c}{Top-1} & \multicolumn{1}{c}{Top-5} &       &       & (GPU days)  \\
    \hline
    ResNet-18~\cite{resnet} & 69.8 & 89.1 & 11.7 & 1814 & -- \\
    Inception-v1~\cite{szegedy2015going} & 69.8      & 89.9      &   6.6    &  1448     & -- \\
    MobileNet~\cite{howard2017mobilenets} & 70.6 & 89.5 & 4.2 & 569 & -- \\
    \hline
    NASNet-A~\cite{zoph2018learning} & 74.0 & 91.6 & 5.3 & 564 & 1800 \\

    NASNet-B~\cite{zoph2018learning} & 72.8 & 91.3 & 5.3 & 488 & 1800 \\
    % NASNet-C~\cite{zoph2018learning} & 27.5 & 9.0 & 4.9 & 558 & 1800 \\
    NASNet-C~\cite{zoph2018learning} & 72.5 & 91.0 & 4.9 & 558 & 1800 \\
    % AmoebaNet-A~\cite{real2018regularized} & 25.5 & 8.0 & 5.1 & 555 & 3150 \\
    AmoebaNet-A~\cite{real2018regularized} & 74.5 & 92.0 & 5.1 & 555 & 3150 \\
    % AmoebaNet-B~\cite{real2018regularized} & 27.2 & 8.7 & 5.3 & 555 & 3150 \\
    AmoebaNet-B~\cite{real2018regularized} & 74.0 & 92.4 & 5.3 & 555 & 3150 \\
    % AmoebaNet-C~\cite{real2018regularized} & 27.5 & 9.0 & 4.9 & 570 & 3150 \\
    % AmoebaNet-C~\cite{real2018regularized} & 75.7 & 91.0 & 4.9 & 570 & 3150 \\
    % PNAS~\cite{liu2018progressive} & 25.8 & 8.1 & 5.1 & 588 & $\sim$255 \\
    GHN~\cite{zhang2018graph} & 73.0 & 91.3 & 6.1 & 569 & 0.8 \\
    % SNAS~\cite{xie2018snas} & 27.3 & 9.2 & 4.3 & 522 & 1.5 \\
    SNAS~\cite{xie2018snas} & 72.7 & 90.8 & 4.3 & 522 & 1.5 \\
    % DARTS~\cite{liu2018darts} & 26.9 & 9.0 & 4.9 & 595 & 4 \\
    DARTS~\cite{liu2018darts} & 73.1 & 91.0 & 4.9 & 595 & 4 \\
    NAT-DARTS~\cite{guo2019nat} & 73.7 & 91.4 & 4.0 & 441 & - \\
    % GHN~\cite{zhang2018graph} & 27.0 & 8.7 & 6.1 & 569 & 0.8 \\
    PNAS~\cite{liu2018progressive} & 73.5 & 91.4 & 5.1 & 588 & 255 \\
    MnasNet-92~\cite{DBLP:conf/cvpr/TanCPVSHL19} & 74.8 & 92.0 & 4.4 & - & - \\
    % SNAS & 72.7 & 90.8 & 4.3 & 522 & 1.5 \\
    ProxylessNAS\cite{cai2018proxylessnas} & 75.1 & 92.5 & 7.1 & - & 8.3 \\
    CARS~\cite{yang2019cars} &75.2 & 92.5 & 5.1 & 591 & 0.4 \\
    % XNAS~\cite{nayman2019xnas}$^\dagger$ & 73.4 & 91.3 & 5.2 & 544 & 0.3\\
    \hline
    \cnas & {\textbf{75.4}} & {\textbf{92.6}} & 5.3 & 576 & \textbf{0.3} \\
    \hline
    \end{tabular}%
    }
\end{table*}

\subsection{Demonstration of \cnas}\label{exp:node_vs_operation}
    
To investigate our \cnas, we compare the performance of the architectures searched in each stage with \fixnas and \cnasnode.
For a fair comparison, {we train  the controller on CIFAR-10 using these three methods for the same epochs, \ie, $320$ epochs in total}. We use the same operation order for both \cnas and \fixnas (\ie, \cnas {and} \fixnas have the same search space in each stage).
We sample architectures at the end of each stage and train them to convergence.
All the architectures are limited to 3.8M parameters in the evaluation.
We show the comparisons of different methods in different {stages} in Figure~\ref{fig:curriculum}.

From Figure \ref{fig:curriculum}, both \cnas and \fixnas architectures obtain better performance as the search space increases.
However, our \cnas architectures consistently outperform \fixnas ones for all stages. 
This implies that directly searching in a large search space (\ie, \fixnas) is more difficult than searching in a progressively growing search space (\ie, \cnas).
Since our \cnas learns the controller in a progressive manner, the knowledge learned in the smaller search space will be transferred to the next training stage.
With the help of knowledge inherited from the previous learning,  \cnas finds better architectures than \fixnas.

Compared with \cnasnode, the architectures found by \cnas achieve better performance at the same epoch.
As for the largest search space, the searched architecture of \cnas also yields significantly better performance than \cnasnode one (97.40\% vs. 97.15\%). 
Moreover, {the improvement of performance} between the last two {stages} in \cnasnode {becomes} smaller.
The reason is {that} the search space is increasing more quickly with the addition of nodes (See Figure~\ref{fig:space_size}), which {introduces} a large gap between the last two stages.
In contrast, the growth of the search space is more smooth with the addition of the operation, the gap between two adjacent stages of our proposed \cnas is smaller than \cnasnode.
As a result, \cnas {finds} better architectures than \cnasnode.

\subsection{Evaluation on CIFAR-10}\label{exp:evaluation_cifar10}
We first search for the convolution cells with our proposed \cnas on CIFAR-10 data set.
Then, we build the final convolution networks by stacking the learned cells and evaluate them on CIFAR-10 data set.

\textbf{Training details}.
{Following} the setting in \cite{liu2018darts}, convolution cells have two types, namely the normal cell and the reduction cell.
Each cell contains 7 nodes, including 2 input nodes, 4 intermediate nodes, and 1 output node. 
The available operations between two nodes include $3 \times 3$ depthwise separable convolution, $5 \times 5$ depthwise separable convolution, $3 \times 3$ max pooling, $3 \times 3$ average pooling, $3 \times 3$ dilated convolution, $5 \times 5$ dilated convolution, identity and none.
We force the first added operation to have parameters (\eg, convolution) for the reason that the sampled network without parameters cannot be trained. 
We divide the official training set of CIFAR-10 {into} two parts, 40\% for training the super network parameters and 60\% for training the controller parameters.
We train the controller for $320$ epochs in total, {with} $40$ epochs for each stage.
{Before adding operations at each stage, we perform the operation warmup for $20$ epochs.}
More details can be found in the supplementary.

\textbf{Evaluation details}.
The final convolution network is stacked with 20 learned cells: 18 normal cells and 2 reduction cells. 
We set the initial number of the channels to 36.
Following~\cite{liu2018darts}, we train the network for $600$ epochs using the batch size of $96$.
We use an SGD optimizer with a weight decay of $3 \times 10^{-4}$ and a momentum of $0.9$.
The learning rate starts from $0.025$ and follows the cosine annealing strategy to a minimum of $0.001$.
We use cutout~\cite{devries2017improved} with a length of 16 for data augmentation.
We report the mean and standard deviation of 10 independent experiments for our final convolution network. More details can be found in the supplementary.

\textbf{Comparisons with state-of-the-art methods}..
We compare our \cnas with state-of-the-art methods in Table~\ref{tab:cifar-10} and show the learned normal and reduction cells in Figure~\ref{fig:cnn_arch}.
The architecture found by \cnas achieves the average test accuracy of $97.40\%$, which outperforms all the considered methods.
By searching in the progressively growing search space, our \cnas makes use of the knowledge inherited rather than train from scratch.
In this way, the architecture search problem becomes simpler.
As a result, \cnas finds better architectures than other methods.
	
\subsection{Evaluation on ImageNet}\label{exp:evaluation_imagenet}

To verify the transferability of the learned cells on CIFAR-10, we evaluate them on a large-scale image classification data set ImageNet, which contains 1,000 classes with 128k training images and 50k testing images. 

\textbf{Evaluation details}.
We stack 14 cells searched on CIFAR-10 to build the final convolution network, with 12 normal cells and 2 reduction cells.
The initial number of the channels is set to 48.
Following the settings in~\cite{liu2018darts}, the network is trained for $250$ epochs with a batch size of $256$.
We use an SGD optimizer with a weight decay of $3 \times 10^{-5}$. The momentum term is set to $0.9$.
% The learning rate is initialized to $0.1$ and annealed via cosine decay to $0.001$.
The learning rate is initialized to $0.1$ and we gradually decrease it to zero.
Following the setting in ~\cite{pham2018efficient,liu2018progressive,liu2018darts}, we consider the mobile setting where multiply-adds (Madds) is restricted to be less than 600M. More details can be found in the supplementary.

\textbf{Comparisons with state-of-the-art methods}.
We compare the performance of the architecture found by \cnas with several state-of-the-art models {and report the results} in Table~\ref{tab:imagenet}.
Under the mobile setting,
the architecture found by \cnas achieves $75.4\%$ top-1 accuracy and $92.6\%$ top-5 accuracy, outperforming the human-designed architectures and NAS based architectures.
Moreover, compared {with} NASNet-A, AmoebaNet-A, and PNAS, our \cnas architecture also achieves competitive performance even with two or three orders of magnitude fewer computation resources.
Compared with other {heavyweight} model, \eg, ResNet-18 and Inception-v1, our model yields better performance with significantly less computation cost.

\section{Further Experiments}

To verify the robustness of the proposed \cnas, we conduct two further experiments to investigate the effect of operation warmup and different operation orders.
    
\subsection{Effect of Operation Warmup}\label{exp:warmup}
We investigate the effect of operation warmup on the search performance of \cnas.
For a fair comparison, we train different controllers with the same number of epochs.
From Figure~\ref{fig:curriculum}, without operation warmup, the controller tends to find sub-optimal architectures and the search performance is also very unstable during the training phase. 
When equipped with the proposed operation warmup, the resultant controller consistently outperforms that without operation warmup in all training stages.
These results demonstrate the necessity and effectiveness of the proposed operation warmup.

\subsection{Effect of Different Operation Orders}\label{exp:order}
We compare the performance of the architectures searched by \cnas with different operation orders.
Since the search space is gradually enlarged by adding operations, different operation orders may correspond to different search spaces, leading to different searched architectures.
We repeat the search experiment 5 times with the same settings except for the orders of adding operations on CIFAR-10.
We report the mean accuracy of these architectures found by \cnas over 5 runs in Figure~\ref{fig:curriculum}.
\cnas achieves better mean accuracy than \fixnas with different operation orders.
The experimental results indicate the proposed \cnas is not sensitive to the orders of the operations.

\section{Conclusion}
In this paper, we have proposed a Curriculum Neural Architecture Search (\cnas) method to alleviate the training difficulty incurred by {the space explosion issue}.
To this end, we propose a curriculum search strategy that solves a series of NAS problems with the search spaces with increasing size and gradually incorporates the learned knowledge to guide the search for a more difficult NAS problem.
To construct these problems, we  gradually introduce new operations into the search space.
By inheriting the knowledge learned from the smaller search spaces, \cnas can greatly improve the search performance in the largest space.
Extensive experiments on CIFAR-10 and ImageNet demonstrate the superiority of \cnas over existing methods.

\section*{Acknowledgements}
This work is partially supported by
Key-Area Research and Development Program of Guangdong Province (No. 2018B010107001, 2019B010155002, 2019B010155001),
National Natural Science Foundation of China (NSFC) 61836003 (key project), 2017ZT07X183, 
the Guangdong Basic and Applied Basic Research Foundation (No. 2019B1515130001), 
the Guangdong Special Branch Plans Young Talent with Scientific and Technological Innovation (No. 2016TQ03X445), 
the Guangzhou Science and Technology Planning Project (No. 201904010197),
Natural Science Foundation of Guangdong Province (No. 2016A030313437), 
Tencent AI Lab Rhino-Bird Focused Research Program (No. JR201902) 
and Fundamental Research Funds for the Central Universities D2191240.

\bibliography{cnas}
\bibliographystyle{icml2020}

\onecolumn
\appendix

\begin{center}
	{
		\Large{\textbf{Supplementary Materials for ``Breaking the Curse of Space Explosion:Towards Efficient NAS with Curriculum Search''}}
	}
\end{center}

We organize our supplementary material as follows.
First, we give more discussions on Eqn.~(3) that calculates the size of search space in Section~\ref{supp:search_space}.
Second, we detail the training method of CNAS in Section~\ref{supp:train_details}.
Third, we provide more evaluation details for the inferred architectures in Section~\ref{supp:evaluation_details}.

\section{More Discussions on Search Space Size}\label{supp:search_space}

In this section, we focus on the cell-based architecture and analyze the effect of the number of nodes $N$ and the number of candidate operations $K$ on the size of search space.

In this paper, we focus on searching for the optimal cell-based architecture~\cite{pham2018efficient,liu2018darts}.
A cell-based architecture can be represented by a directed acyclic graph (DAG), \ie, $\alpha = (\mathcal{V}, \mathcal{E})$. 
$\mathcal{V}$ is the set of the nodes that represent the feature maps in the neural networks.
$\mathcal{E}$ is the set of the edges that represent some computational operations (\eg, convolution or max pooling). 
For convenience, we denote the number of nodes and candidate operations by $B$ and $K$, respectively. In the following, we will depict the design of cell-based architectures.
    
Following~\cite{pham2018efficient,liu2018darts}, a cell-based architecture consists of two input nodes, $B-3$ intermediate nodes, and one output {node}.
The input nodes denote {the outputs of} the nearest two cells in front of the current one. 
The output node {concatenates} the outputs of all the intermediates to produce the final output of the cell. 
In the DAG, each intermediate node also takes two previous nodes in this cell as inputs.
In this sense, there are $2 \times (B - 3)$ edges in the DAG and we will determine which operation should be applied out of $K$ candidate operations.
Based on the design of cell-based architecture, we will analyze the effect of both the number of nodes $N$ and the number of candidate operations $K$ on the size of search space.

First, we calculate the number of cell-based topological structures while ignoring the type of edges (\ie, operations).
Since each intermediate node must take two previous nodes in the same cell as inputs, there are $i-1$ optional previous nodes for the $i$-th intermediate node ($3 \leq i \leq B-1$).
In this case, the $i$-th node has $(i-1)^2$ choices for the two inputs.
To sum up, the number of the cell-based topological structure, denoted by $M$, can be calculated by
\begin{equation}
    M = \prod_{i=3}^{B-1} (i-1)^2 = \big((B-2)!\big)^2.  \tag{A}
\end{equation} 

Then, we calculate the number of all the possible variants w.r.t a specific topological structure.
For a specific topological structure, there exist many variants because the edges of each intermediate node have not determined which operation to choose. 
Since the cell has $B - 3$ intermediate nodes and each intermediate node has two inputs from two previous nodes, 
there are $2 \times (B - 3)$ edges that we should decide which operation should be chosen. 
Moreover, each edge has $K$ operations to be chosen. 
Because the choices are independent, the number of all the possible variants w.r.t a specific topological structure, denoted by $L$, can be calculated by
\begin{equation}
    L = K^{2(B-3)}.  \tag{B}
\end{equation} 

In conclusion, given $B$ nodes and $K$ candidate operations, the size of the search space $\Omega$ becomes: 
\begin{equation}\label{supp:eq:space_size}
    |\Omega| = M \times L = K^{2(B-3)}\big((B-2)!\big)^2. \tag{C}
\end{equation}
    
From Eqn.~(\ref{supp:eq:space_size}), the search space can be extremely large when we have a large $B$ or $K$.
For example, $|\Omega| \approx 5 \times 10^{12}$ in ENAS~\cite{pham2018efficient} when $B = 8$ and $K = 5$, $|\Omega| \approx 2 \times 10^{11}$ in DARTS~\cite{liu2018darts} when $B = 7$ and $K = 8$.
Since the size of the search space is extremely large, searching for the optimal architectures in such a large search space is a difficult problem.

\section{More Training Details of CNAS}\label{supp:train_details}

Following the settings in \cite{liu2018darts}, the convolutional cells have two types, namely the normal cell and the reduction cell.
The normal cell keeps the spatial resolution of the feature maps.
The reduction cell reduces the height and width of the feature maps by half and doubles the number of the channels of the feature maps.
Each cell contains 7 nodes, including 2 input nodes, 4 intermediate nodes, and 1 output node.
The output node is the concatenation of the 4 intermediate nodes.
There are 8 candidate operations between two nodes, including $3 \times 3$ depthwise separable convolution, $5 \times 5$ depthwise separable convolution, $3 \times 3$ max pooling, $3 \times 3$ average pooling, $3 \times 3$ dilated convolution, $5 \times 5$ dilated convolution, identity, and none.
The convolutions are applied in the order of ReLU-Conv-BN.
Each depthwise separable convolution is applied twice following \cite{zoph2018learning}.
    
We search for good architectures on CIFAR-10 and evaluate the learned architectures on both CIFAR-10 and ImageNet.
Note that some of the operations do not contain any parameters. Thus, a model cannot be trained if a cell-based architecture only consists of the operations without parameters.
To avoid this issue, we make the first operation in the operation sequence to have trainable parameters, \eg, depthwise separable convolution or dilated convolution.

In the training, we divide the original training set of CIFAR-10 into two parts, 40\% for training the super network parameters and 60\% for training the controller parameters~\cite{pham2018efficient}.
We train the controller for $320$ epochs in total, $40$ epochs for each stage.
Before adding operations at each stage, we perform the operation warmup for $20$ epochs.
For training the super network, we use SGD optimizer with a weight decay of $3 \times 10^{-4}$ and a momentum of $0.9$. The learning rate is set to $0.1$.
For training the controller, we use ADAM with a learning rate of $3 \times 10^{-4}$ and a weight decay of $5 \times 10^{-4}$.
We add the controller's sample entropy to the reward, which is weighted by $0.005$.

\section{More Evaluation Details on CIFAR-10 and ImageNet}\label{supp:evaluation_details}
    
In this section, we provide more details about the evaluation method. Following~\cite{liu2018darts}, we conduct architecture search on CIFAR-10 and evaluate the searched architectures on two benchmark datasets, namely CIFAR-10 and ImageNet. 
    
\noindent \textbf{More evaluation details on CIFAR-10.}
We build the final convolution network with 20 learned cells, including 18 normal cells and 2 reduction cells.
The two reduction cells are put at the $1/3$ and $2/3$ depth of the network, respectively.
The initial number of the channels is set to 36.
Following the setting in~\cite{liu2018darts}, we train the convolution network for $600$ epochs using the batch size of $96$.
The training images are padded 4 pixels on each side.
Then the padded images or its horizontal flips are randomly cropped to the size of $32 \times 32$.
We also use cutout~\cite{devries2017improved} with a length of $16$ in the data augmentation.
We use the SGD optimizer with a weight decay of $3 \times 10^{-4}$ and a momentum of $0.9$.
The learning rate starts from $0.025$ and follows the cosine annealing strategy with a minimum of $0.001$.
Additional enhancements include path dropout of the probability of $0.2$ and auxiliary towers~\cite{szegedy2015going} with a weight of $0.4$.

\textbf{More evaluation details on ImageNet.}
We consider the \textbf{\emph{mobile setting}} to evaluate our ImageNet models. In the mobile setting, the input image size is $224\times224$ and the number of multiply-adds (MAdds) in the model is restricted to be less than 600M.
We build the network with 11 cells, including 9 normal cells and 2 reduction cells.
The locations of the two reduction cells are the same as the setting of CIFAR-10 experiments.
We follow the same setting as that in DARTS~\cite{liu2018darts}.
Specifically, we perform the same data augmentations during training, including horizontally flipping, random crops, and color jittering.
During testing, we resize the input images to $256 \times 256$ and then apply a center crop to the size $224 \times 224$.
The network is trained for $250$ epochs with a batch size of $256$.
We use SGD optimizer with a weight decay of $3 \times 10^{-5}$ and a momentum of $0.9$.
The learning rate is initialized to $0.1$ and we gradually decrease it to zero.
%The learning rate is initialized to $0.1$ and decreases using the cosine annealing with a minimum of $0.001$.
We add auxiliary loss with a weight of $0.4$ after the last reduction cell.
We set the path dropout to the probability of $0.2$ and label smoothing to $0.1$.

\end{document}